\documentclass[10pt,twocolumn,letterpaper]{article}

\usepackage[pagenumbers]{cvpr} 

\usepackage{pifont}
\newcommand{\cmark}{\ding{51}}
\newcommand{\xmark}{\ding{55}}

\usepackage{multirow}

\usepackage{pgfplots}
\pgfplotsset{compat=1.17}

\definecolor{cvprblue}{rgb}{0.21,0.49,0.74}
\usepackage[pagebackref,breaklinks,colorlinks,allcolors=cvprblue]{hyperref}

\def\confName{CVPR}
\def\confYear{2027}

\title{GeoMVGen: Joint Appearance-Geometry Modeling for Multi-View Video Generation in Robotic Manipulation}

\author{%
Qiaoqiao Jia\textsuperscript{1,2,*} \quad
Guile Wu\textsuperscript{1} \quad
Dongfeng Bai\textsuperscript{1}\\
\textsuperscript{1}Huawei Noah's Ark Lab \quad
\textsuperscript{2}University of Waterloo\\
{\tt\small christinajia0803@gmail.com\quad guile.wu@outlook.com\quad baidongfeng@huawei.com}
}

\begin{document}

\twocolumn[{%
\renewcommand\twocolumn[1][]{#1}%
\maketitle
\begin{center}
    \centering
    \captionsetup{type=figure}
    \includegraphics[
  width=0.95\textwidth,
]{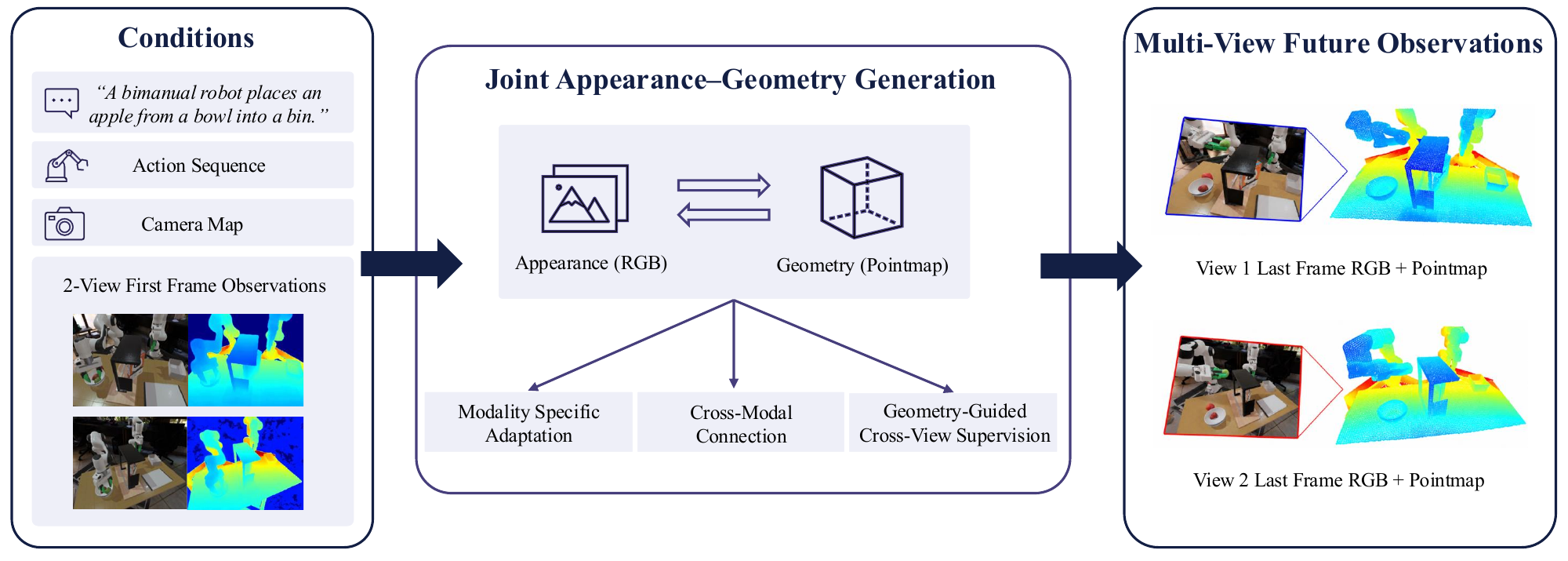}
    \captionof{figure}{
        GeoMVGen overview.
        Given input conditions, GeoMVGen jointly generates future multi-view RGB videos and corresponding pointmap sequences, achieving realistic appearance and geometric consistency across viewpoints.
        }
    \label{fig:main}
    \vspace{6pt}
\end{center}%
}]

{\renewcommand{\thefootnote}{*}%
\footnotetext{Qiaoqiao Jia contributed to this work during an internship at Huawei Canada.}}

\begin{abstract}
Robotic manipulation requires predicting how robot-object interactions evolve over time, encompassing both visual appearance and scene geometry.
In multi-camera settings, such predictions need to capture the evolving 3D scene consistently across viewpoints, while supporting controllable generation under different robot actions and camera configurations.
Existing video generation models provide strong priors for visual appearance and motion but typically lack explicit geometric modeling.
Meanwhile, recent advances in geometry-aware and multi-view generation have explored joint appearance-geometry modeling and cross-view consistency, yet generating temporally coherent appearance and geometry across multiple viewpoints under robotic manipulation conditions remains challenging.
In this work, we propose GeoMVGen, a framework for joint multi-view RGB and pointmap video generation in robotic manipulation.
Built on a video diffusion transformer, our framework employs modality-specific adaptation modules to capture the distinct characteristics of appearance and geometry, while using cross-modal residual connections to facilitate information exchange between the two modalities.
To establish geometric correspondences across viewpoints, we further introduce a cross-view projection mechanism with complementary geometric supervisions, combining pointmap consistency with geometry foundation model-guided feature alignment.
These designs enable controllable generation of temporally evolving multi-view observations with coherent appearance and geometrically consistent scene structures.
Extensive experiments on 4D video generation dataset comprising simulated and real-world robotic manipulation scenarios demonstrate that GeoMVGen outperforms state-of-the-art methods in video quality, geometric accuracy, and cross-view consistency.
\end{abstract}    
\section{Introduction}
\label{sec:intro}

Predicting how robot-object interactions evolve during robotic manipulation requires modeling changes in both visual appearance and 3D scene geometry.
In multi-camera settings, generating future observations further requires temporal coherence within each view and geometric consistency across viewpoints.
Moreover, different robot actions and camera configurations introduce additional demands for controllable generation under flexible conditioning inputs.
These requirements motivate joint multi-view appearance-geometry generation that preserves visual realism and geometric consistency during robotic manipulation.

Recent advances in general-purpose video generation~\cite{blattmann2023stable,yang2025cogvideox,wan2025wan,chen2024videocrafter2,bar2024lumiere,kong2024hunyuanvideo} have established strong priors for visual appearance and motion, providing a foundation for video generation in robotic manipulation~\cite{du2023learning,hu2024video,liang2025video,bharadhwaj2024gen2act}.
However, visually plausible predictions do not necessarily capture accurate scene geometry.
To address this limitation, methods such as TesserAct~\cite{zhen2025tesseract} and 4DNeX~\cite{chen20254dnex} explore joint appearance-geometry generation, while One4D~\cite{mi2026one4d} and UniVidX~\cite{chen2026unividx} further investigate modality-specific adaptation and cross-modal information exchange for multimodal generation.
Meanwhile, multi-view video generation methods~\cite{xie2025sv4d,wu2025cat4d,kuang2024collaborative,bai2025syncammaster} have advanced the generation of consistent videos across multiple viewpoints.
More recently, 4DGen~\cite{liu2026geometry} and MVISTA-4D~\cite{wang2026mvista} have extended multi-view generation to jointly model appearance and geometry for robotic manipulation.
Despite these advances, effectively integrating modality-specific representation learning, cross-modal information exchange, and cross-view geometric consistency while maintaining high-quality video generation remains challenging in robotic manipulation.

In this work, we propose GeoMVGen, a new framework for joint multi-view RGB and pointmap video generation in robotic manipulation.
Built on a video diffusion transformer, GeoMVGen employs modality-specific adaptation modules to capture the distinct characteristics of appearance and geometry representations.
Furthermore, GeoMVGen uses bidirectional cross-modal residual connections to facilitate information exchange between the two modalities, while preserving their separate adaptation pathways.
To encourage geometric consistency across viewpoints, GeoMVGen establishes cross-view correspondences through geometric projection and camera calibration, and introduces two complementary supervision objectives.
These two objectives serve complementary roles: cross-view pointmap consistency supervision directly constrains corresponding 3D predictions, while geometry-aware cross-view feature alignment supervision encourages consistency between representations extracted from generated RGB frames using a geometry foundation model~\cite{wang2026vggt}.
With different input conditions, such as initial RGB observations and pointmaps from multiple viewpoints, language instructions, and optional robot action sequences and camera parameters, GeoMVGen generates future appearance and geometry in a controllable and geometrically consistent manner.
We evaluate GeoMVGen on both simulated and real-world robotic manipulation data.
Extensive experiments show the superiority of GeoMVGen over state-of-the-art methods in video quality, geometric accuracy, and cross-view consistency.

Our contributions are summarized as follows:
\begin{itemize}
    \item We propose GeoMVGen, a new framework for controllable multi-view appearance and pointmap video generation in robotic manipulation.
    \item We equip GeoMVGen with modality-specific adaptation and bidirectional cross-modal residual connections to capture the distinct characteristics of appearance and geometry representations while enabling effective information exchange between the two modalities.
    \item We design GeoMVGen with a cross-view consistency mechanism that integrates complementary geometric supervision based on pointmap consistency and geometry foundation model-guided feature alignment.
\end{itemize}

\begin{figure*}[t]
\centering
\includegraphics[width=0.95\textwidth]{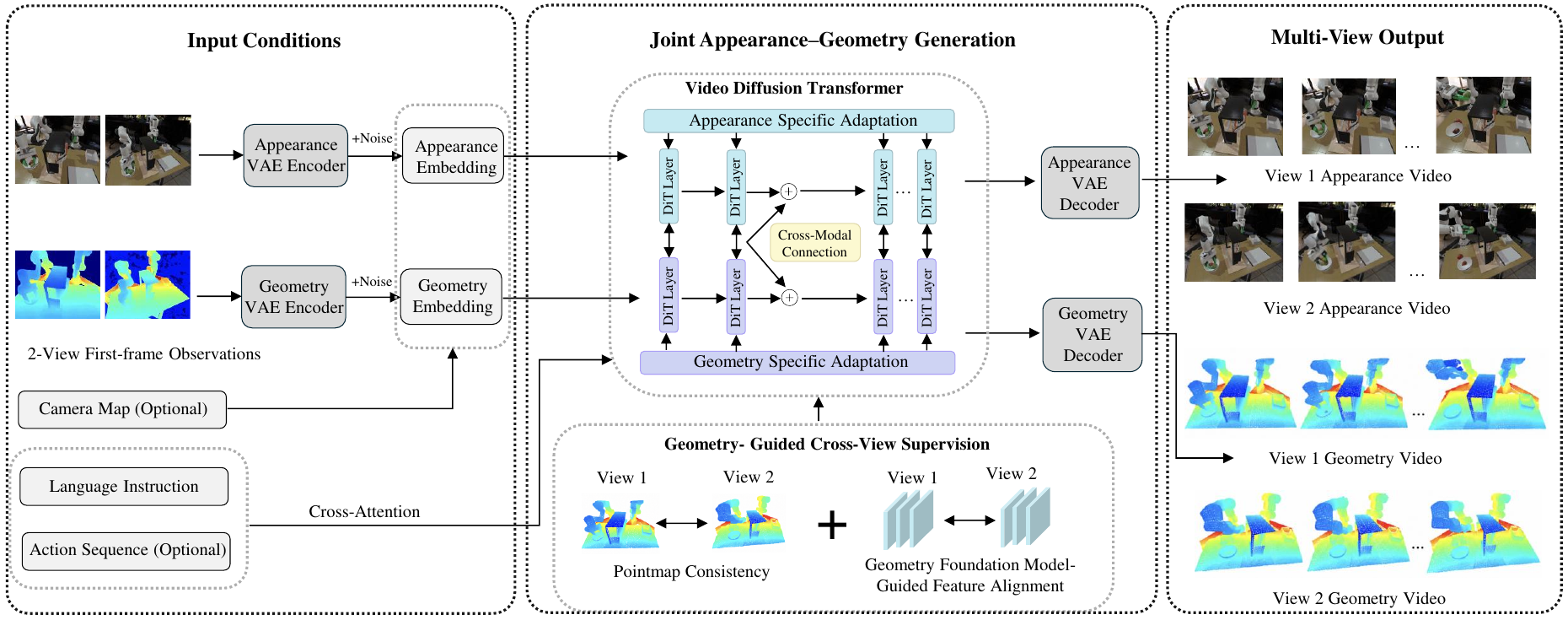}
\caption{Overall framework of GeoMVGen.
Given input conditions, GeoMVGen generates future multi-view RGB videos and pointmaps in a controllable and geometrically consistent manner.
GeoMVGen is built on a video diffusion transformer with modality-specific adaptation, bidirectional cross-modal residual connections, and a cross-view consistency mechanism with complementary geometric supervision.}
\label{fig:framework}
\end{figure*}
\section{Related Work}
\label{sec:related_work}

\paragraph{Video Generation for Robotic Manipulation.}
General-purpose video generation models \cite{blattmann2023stable,yang2025cogvideox,wan2025wan,chen2024videocrafter2,bar2024lumiere,kong2024hunyuanvideo,liang2025video} have learned rich appearance and motion priors, providing a foundation for video generation in robotic manipulation.
Building on these advances, recent approaches have explored video generation for various purposes, such as visual planning, predictive representation learning, and robot policy learning.
UniPi \cite{du2023learning} generates text-conditioned visual plans and converts them into executable actions through inverse dynamics, while Video Prediction Policy \cite{hu2024video} leverages predictive representations from video diffusion models to guide robot policies.
These approaches primarily focus on planning, predictive representations, or policy learning.
In contrast, our work focuses on generating future appearance and geometry observations with explicit geometric consistency across viewpoints in robotic manipulation.

\vspace{-0.3cm}
\paragraph{Joint Appearance-Geometry Modeling for Video Generation.}
Joint modeling of appearance and geometry enables video generators to capture both visual dynamics and the underlying 3D scene structure~\cite{zhen2025tesseract,mi2026one4d,chen2026unividx,Wu_2026_CVPR}.
TesserAct \cite{zhen2025tesseract} jointly predicts RGB, depth, and normal videos, followed by spatial and temporal geometry refinement, while 4DNeX \cite{chen20254dnex} jointly models RGB and XYZ sequences to generate dynamic point clouds from a single image.
Geometry Forcing \cite{wu2026geometry} aligns intermediate diffusion features with pretrained VGGT \cite{wang2025vggt} representations through angular and scale-based objectives, while ViCoDR \cite{danier2025view} uses 3D correspondences to supervise internal diffusion features for view-consistent generation.
However, they do not specifically address joint multi-view appearance-geometry generation for robotic manipulation.
For multi-view joint appearance-geometry generation in robotic manipulation, 4DGen \cite{liu2026geometry} jointly generates multi-view RGB videos and pointmaps, promoting geometric consistency through pointmap alignment supervision and cross-view attention in its pointmap prediction branches.
MVISTA-4D \cite{wang2026mvista} combines cross-modal fusion and geometry-aware cross-view attention for multi-view RGB-D generation, supporting test-time action inference through trajectory-latent optimization.
Although these methods show promising results, multi-view appearance-geometry generation through cross-view modeling and cross-modal interaction remains challenging.
Our framework combines modality-specific adaptation and cross-modal residual connections with complementary output-level supervision, directly constraining predicted 3D geometry while aligning geometry-aware visual features across viewpoints.

\section{Methodology}

\subsection{The Overall Framework}

Our work focuses on multi-view (multi-camera) appearance and geometry joint modeling for video generation in robotic manipulation.
Let $\mathcal{O}_0=\{(I_0^v,P_0^v)\}_{v=1}^{2}$ denote the initial RGB observations and pointmaps from two viewpoints.
Conditioned on $\mathcal{O}_0$, a language instruction $l$, and optional robot action sequences $\mathbf{a}$ and camera parameters $\mathbf{c}$, we model the joint distribution of future RGB and pointmap sequences as
\begin{equation}
p_{\theta}\!\left(
\{(I_{1:T}^{v},P_{1:T}^{v})\}_{v=1}^{2}
\mid \mathcal{O}_0,l,{\mathbf{a}},{\mathbf{c}}
\right),
\end{equation}
where $I_{1:T}^{v}$ and $P_{1:T}^{v}$ denote the future RGB and pointmap sequences of view $v$, respectively.
As illustrated in Fig.~\ref{fig:framework}, GeoMVGen addresses three key aspects of multi-view appearance-geometry generation, namely joint modeling of distinct modalities, cross-view geometric consistency, and flexible conditioning.
RGB and pointmaps are encoded into separate latent streams and processed using a shared video diffusion transformer~\cite{wan2025wan}.
Modality-specific adaptation captures the characteristics of each representation, while bidirectional cross-modal residual connections enable information exchange between the two streams.
To promote geometric consistency across viewpoints, GeoMVGen incorporates a cross-view consistency mechanism with two complementary supervision objectives: pointmap consistency and geometry-aware feature alignment.
Language instructions and optional robot actions guide generation through cross-attention, while camera conditioning supplies spatially aligned viewpoint information.

\subsection{Joint Appearance-Geometry Modeling}

RGB and pointmaps provide complementary representations of visual appearance and 3D scene structure.
However, their distinct characteristics require specialized representation learning, while joint generation also benefits from information exchange between the two modalities.
To achieve this, we employ LoRA~\cite{hu2021lora} based modality-specific adaptation and bidirectional cross-modal residual connections within a shared video diffusion backbone.

\vspace{-0.3cm}
\paragraph{Modality-Specific Adaptation.}
For a linear transformation with pretrained weight $W \in \mathbb{R}^{d_{\mathrm{out}}\times d_{\mathrm{in}}}$ and bias $b$, the output for modality $m$ is
\begin{equation}
    f_m(h)
    = Wh+b+\frac{\alpha}{r}B_m A_m h,
    \quad m\in\{\mathrm{rgb},\mathrm{xyz}\},
\end{equation}
where $A_m\in\mathbb{R}^{r\times d_{\mathrm{in}}}$ and $B_m\in\mathbb{R}^{d_{\mathrm{out}}\times r}$ are trainable low-rank matrices, $r$ is the adapter rank, and $\alpha$ controls the scaling of the adaptation.
The pretrained weights are frozen and shared across modalities, while the adapter parameters are modality-specific and shared across camera views.
We apply these adapters to all linear layers of the backbone, with $r=\alpha=64$.
Each $B_m$ is initialized to zero, so the low-rank updates initially preserve the pretrained transformations.

\vspace{-0.3cm}
\paragraph{Cross-Modal Residual Connections.}
While modality-specific adaptation preserves the characteristics of each modality, independent processing alone limits their ability to exchange complementary information.
We therefore introduce bidirectional cross-modal residual connections after selected DiT blocks.
Let $h_{\mathrm{rgb},v,k}^{\ell}$ and $h_{\mathrm{xyz},v,k}^{\ell}$ denote the RGB and pointmap features at spatiotemporal token $k$ in view $v$, obtained from block $\ell$.
At each selected block, both streams are updated using their original features as
\begin{equation}
\begin{aligned}
    \tilde{h}_{\mathrm{rgb},v,k}^{\ell}
    &= h_{\mathrm{rgb},v,k}^{\ell}
    +C_{\mathrm{xyz}\rightarrow\mathrm{rgb}}^{\ell}
     h_{\mathrm{xyz},v,k}^{\ell}, \\
    \tilde{h}_{\mathrm{xyz},v,k}^{\ell}
    &= h_{\mathrm{xyz},v,k}^{\ell}
    +C_{\mathrm{rgb}\rightarrow\mathrm{xyz}}^{\ell}
     h_{\mathrm{rgb},v,k}^{\ell}.
\end{aligned}
\end{equation}
Each $C^{\ell}$ is a trainable bias-free linear projection shared across tokens and views.
Both projections are initialized to zero, allowing cross-modal interaction to be learned gradually.
Together, modality-specific adaptation and cross-modal residual connections enable specialized representation learning and effective appearance-geometry information exchange.

\subsection{Cross-View Consistency Learning}

Although joint appearance-geometry modeling enables multi-modal learning, it does not directly enforce consistency between different camera views.
We therefore design a cross-view consistency mechanism with two complementary supervision objectives.
Cross-view pointmap consistency constrains corresponding predicted 3D coordinates, while geometry-aware feature alignment encourages agreement between visual representations at geometrically corresponding locations.
Both objectives use correspondences established through geometric projection with ground-truth geometry and camera calibration during training.

\vspace{-0.3cm}
\paragraph{Cross-View Pointmap Consistency Supervision.}
We enforce agreement between predicted pointmaps expressed in a shared 3D coordinate frame.
For each future frame, we consider both ordered view pairs $(i,j)\in\{(1,2),(2,1)\}$ and omit the temporal index for clarity.
Let $\hat{\mathbf{P}}_i$ denote the predicted pointmap in source view $i$.
Using ground-truth geometry and camera calibration, we project each valid source pixel $\mathbf{u}$ into target view $j$, obtaining its corresponding location $\mathbf{q}_{i\rightarrow j}(\mathbf{u})$.
We then compare the predicted source point with the target point bilinearly sampled from $\hat{\mathbf{P}}_j$ at the projected location.
To account for visibility changes and occlusions during robot-object interactions, we construct a binary validity mask $M_{i\rightarrow j}$ that excludes correspondences with invalid geometry, non-positive depths, or projections outside the target image.
We further compute the cross-view depth discrepancy as
\begin{equation}
    \Delta z_{i\rightarrow j}(\mathbf{u})
    = z_{i\rightarrow j}(\mathbf{u})
    - z_j\!\left(
    \mathbf{q}_{i\rightarrow j}(\mathbf{u})
    \right),
\end{equation}
where $z_{i\rightarrow j}(\mathbf{u})$ is the depth of the source ground-truth point projected into the target camera, and $z_j(\mathbf{q}_{i\rightarrow j}(\mathbf{u}))$ is the target ground-truth depth sampled at the projected location.
For a positive depth tolerance $\tau$, the soft visibility weight is
\begin{equation}
    v_{i\rightarrow j}(\mathbf{u})
    = \exp\left(
    -|\Delta z_{i\rightarrow j}(\mathbf{u})|/\tau
    \right),
\end{equation}
which reduces the contribution of depth-inconsistent correspondences, including those affected by occlusion.
To emphasize robot-object interactions, we also construct foreground masks from available annotations and detected motion.
For each view, motion regions are identified by comparing the ground-truth RGB images and 3D coordinates of future frames against the initial observation, using mean absolute RGB differences and Euclidean coordinate distances, respectively.
We take the temporal union of these regions, apply spatial dilation, and use the resulting motion mask for all frames before combining it with available per-frame foreground annotations by taking their union.
Let $F_i(\mathbf{u})$ denote the resulting binary foreground mask.
We define
\begin{equation}
    g_{i\rightarrow j}(\mathbf{u})
    = 1 + (\gamma-1)\mathbb{I}\Big[
        F_i(\mathbf{u}) \lor F_j\!\left(\mathbf{q}_{i\rightarrow j}(\mathbf{u})
        \right)
    \Big],
\end{equation}
where $\gamma>1$ emphasizes correspondences whose source or target location belongs to the foreground.
The target foreground mask is sampled using nearest-neighbor interpolation.
Combining validity, visibility, and foreground emphasis gives
\begin{equation}
    w^{\mathrm{point}}_{i\rightarrow j}(\mathbf{u})
    = M_{i\rightarrow j}(\mathbf{u})\,
    v_{i\rightarrow j}(\mathbf{u})\,
    g_{i\rightarrow j}(\mathbf{u}).
\end{equation}
We measure the discrepancy between corresponding predicted 3D points using a scene-scale-normalized smooth $\ell_1$ penalty:
\begin{equation}
    e_{i\rightarrow j}(\mathbf{u})
    = \rho\Bigg(
        \frac{
            \hat{\mathbf{P}}_i(\mathbf{u})
            - \hat{\mathbf{P}}_j\!\left(
                \mathbf{q}_{i\rightarrow j}(\mathbf{u})
            \right)
        }{s}
    \Bigg),
\end{equation}
where $\rho$ averages the coordinate-wise smooth $\ell_1$ penalties and $s>0$ is a fixed scene-scale normalization factor.
For each frame and ordered view pair, the loss is
\begin{equation}
    \ell^{\mathrm{point}}_{i\rightarrow j}
    = \frac{
        \sum_{\mathbf{u}}
        w^{\mathrm{point}}_{i\rightarrow j}(\mathbf{u})
        e_{i\rightarrow j}(\mathbf{u})
    }{
        \sum_{\mathbf{u}}
        w^{\mathrm{point}}_{i\rightarrow j}(\mathbf{u})
        + \epsilon
    }.
\end{equation}
We average over valid future frames and both view directions to obtain $\mathcal{L}_{\mathrm{point}}$.
Correspondence coordinates and weighting masks are derived from ground-truth data and detached from gradient computation.
They are used only for supervision and are not required during inference.

\vspace{-0.3cm}
\paragraph{Geometry-Aware Cross-View Feature Alignment.}
Explicit pointmap consistency constrains geometric agreement but does not directly supervise the geometry-aware visual representations of generated RGB frames.
We therefore introduce a complementary feature alignment objective using a frozen VGGT-$\Omega$~\cite{wang2026vggt} encoder.
Its geometry-aware representations provide feature-level constraints complementary to the explicit 3D coordinate alignment enforced by pointmap consistency.

For each decoded RGB prediction $\hat{\mathbf{I}}_i$, the geometry-aware feature map is extracted as $\mathbf{F}_i = \Phi(\hat{\mathbf{I}}_i)$, where $\Phi$ denotes the frozen VGGT-$\Omega$ encoder.
Feature correspondences follow the same ground-truth geometric projection as pointmap consistency, with image coordinates scaled to the resolution of the extracted feature maps.
In the following, $\mathbf{F}_i(\mathbf{u})$ denotes bilinear sampling at the feature-map location corresponding to image coordinate $\mathbf{u}$.
We retain the visibility weight $v_{i\rightarrow j}$, but additionally reject correspondences with $|\Delta z_{i\rightarrow j}(\mathbf{u})|>\tau$ or locations near depth discontinuities in either view.
These filters reduce unreliable feature comparisons under occlusion and at surface boundaries.
Let $M^{\mathrm{feat}}_{i\rightarrow j}$ denote the resulting binary mask.
We define
\begin{equation}
    w^{\mathrm{feat}}_{i\rightarrow j}(\mathbf{u})
    = M^{\mathrm{feat}}_{i\rightarrow j}(\mathbf{u})\,
    v_{i\rightarrow j}(\mathbf{u}).
\end{equation}
The discrepancy between corresponding features is measured by cosine distance:
\begin{equation}
    d_{i\rightarrow j}(\mathbf{u})
    = 1-\operatorname{cos}\Big(
        \mathbf{F}_i(\mathbf{u}), \mathbf{F}_j(
        \mathbf{q}_{i\rightarrow j}(\mathbf{u}))
    \Big),
\end{equation}
where both feature values are sampled at the corresponding spatial locations.
The pairwise feature alignment loss is
\begin{equation}
    \ell^{\mathrm{feat}}_{i\rightarrow j}
    = \frac{
        \sum_{\mathbf{u}}
        w^{\mathrm{feat}}_{i\rightarrow j}(\mathbf{u})
        d_{i\rightarrow j}(\mathbf{u})
    }{
        \sum_{\mathbf{u}}
        w^{\mathrm{feat}}_{i\rightarrow j}(\mathbf{u})
        + \epsilon
    }.
\end{equation}
Averaging over valid sampled future frames and both view directions yields $\mathcal{L}_{\mathrm{feat}}$.
This objective encourages generated views to agree in a geometry-informed feature space.
The encoder parameters remain fixed, while gradients from the feature loss propagate through the encoder to the generated RGB predictions.
Ground-truth geometry supplies correspondence and visibility supervision rather than target appearance features.

\subsection{Controllable Generation}

To support controllable generation under different robot actions and camera configurations, we incorporate action and camera information through separate conditioning pathways.
Action embeddings are integrated into the existing cross-attention conditioning context to provide global motion guidance, while camera embeddings are concatenated channel-wise with the latent inputs before patch embedding to provide spatially aligned viewpoint information.
Both conditioning pathways are applied to the RGB and pointmap streams.

\vspace{-0.3cm}
\paragraph{Action Conditioning.}
We represent each robot action as a 20-dimensional vector comprising the Cartesian end-effector positions, 6D rotation representations, and gripper control values for both arms.
The action sequence is provided by the dataset and temporally aligned with each video clip.
At inference time, the model conditions on the action sequence over the prediction horizon without accessing future ground-truth RGB frames or pointmaps.
Let $\mathbf{a}=\{a_s\}_{s=1}^{S}$ denote the action sequence aligned with the video clip.
Each normalized action vector $a_s\in\mathbb{R}^{20}$ is encoded using a linear projection, a SiLU activation, and a learned temporal embedding:
\begin{equation}
    e_s^{a}
    = W_{\mathrm{out}}^{a}
    \left[
        \operatorname{SiLU}(W_{\mathrm{in}}^{a}a_s+b_{\mathrm{in}}^{a})
        +p_s
    \right]
    +b_{\mathrm{out}}^{a},
\end{equation}
where $p_s$ is a learned temporal embedding, and $W_{\mathrm{out}}^{a}$ and $b_{\mathrm{out}}^{a}$ define a zero-initialized projection to the text-context dimension.
The action embeddings are zero-padded to a fixed capacity and added to the trailing positions of the existing text context without increasing its sequence length.
This conditioning context is shared across views and injected into both modality streams through the backbone's cross-attention layers.

\begin{table*}[t]
\centering
\caption{Comparison on simulated robotic manipulation tasks. We report the average of each metric across the three simulation tasks.
`--' indicates that the method lacks the outputs required to compute the corresponding metrics.}
\label{tab:sota_comparison}
\small
\setlength{\tabcolsep}{4pt}
\resizebox{0.9\textwidth}{!}{%
\begin{tabular}{l cc cccc cc}
\toprule
\multirow{2}{*}{Method}
& \multicolumn{2}{c}{RGB Video Quality}
& \multicolumn{4}{c}{Depth Accuracy}
& \multicolumn{2}{c}{Cross-View Consistency} \\
\cmidrule(lr){2-3}
\cmidrule(lr){4-7}
\cmidrule(lr){8-9}
& $\mathrm{FVD}_{n}$ $\downarrow$
& $\mathrm{FVD}_{m}$ $\downarrow$
& $\mathrm{Abs\,Rel}_{n}$ $\downarrow$
& $\mathrm{Abs\,Rel}_{m}$ $\downarrow$
& $\delta_{1,n}$ $\uparrow$
& $\delta_{1,m}$ $\uparrow$
& CD $\downarrow$
& mIoU $\uparrow$ \\
\midrule
Wan2.1\cite{wan2025wan}
& 530.0 & 560.6 & -- & -- & -- & -- & -- & -- \\
SVD\cite{blattmann2023stable}
& 739.1 & 728.6 & -- & -- & -- & -- & -- & -- \\
4DGen\cite{liu2026geometry}
& 648.9 & 636.6 & 0.076 & 0.087 & 0.935 & 0.934 & 0.029 & 0.496 \\
TesserAct\cite{zhen2025tesseract}
& 724.4 & 713.1 & 0.159 & 0.149 & 0.897 & 0.907 & -- & -- \\
Geometry Forcing\cite{wu2026geometry}
& 1107.8 & 1094.1 & 0.290 & 0.267 & 0.517 & 0.580 & -- & -- \\
\midrule
\textbf{Ours}
& \textbf{480.2} & \textbf{486.5} & \textbf{0.051} & \textbf{0.049} & \textbf{0.957} & \textbf{0.960} & \textbf{0.013} & \textbf{0.530} \\
\bottomrule
\end{tabular}%
}
\end{table*}
\begin{figure*}[t!]
\centering
\includegraphics[
  width=0.92\textwidth
]{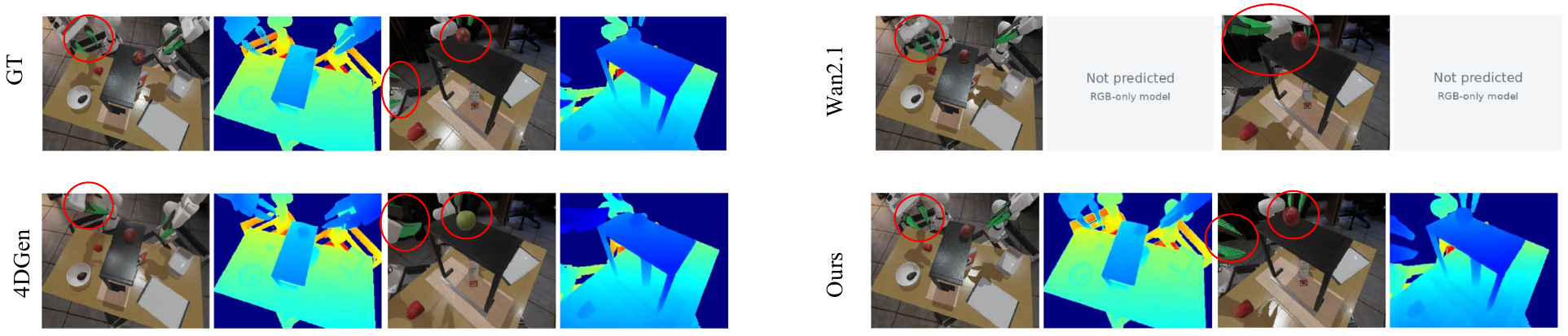}
\caption{Qualitative comparison with state-of-the-art methods on simulated robotic manipulation tasks.}
\label{fig:qualitative}
\end{figure*}

\vspace{-0.3cm}
\paragraph{Camera Conditioning.}
We encode camera parameters as a spatially aligned, nine-channel geometric map containing the ray direction, Pl\"ucker moment, and camera position at each spatial location.
For a pixel with homogeneous coordinates $\tilde{\mathbf{u}}=(u,v,1)^\top$, its unit ray direction is
\begin{equation}
    \mathbf{d}(\mathbf{u}) =
    \frac{\mathbf{R}\mathbf{K}^{-1}\tilde{\mathbf{u}}}
    {\|\mathbf{R}\mathbf{K}^{-1}\tilde{\mathbf{u}}\|_2},
\end{equation}
where $\mathbf{K}$ denotes the camera intrinsics and $\mathbf{R}$ rotates camera-coordinate vectors into the shared world coordinate system.
The direction $\mathbf{d}$ specifies where the viewing ray points.
Given the camera center $\mathbf{o}$ in the same coordinate system, the Pl\"ucker moment $\mathbf{m}(\mathbf{u})=\mathbf{o}\times\mathbf{d}(\mathbf{u})$ encodes the spatial location of the ray relative to the coordinate origin.
Together, $\mathbf{d}$ and $\mathbf{m}$ represent the oriented viewing line, while $\mathbf{o}$ explicitly specifies the camera center along that line.
We construct the camera encoding as
\begin{equation}
    \mathbf{c}(\mathbf{u}) =
    \left[
    \mathbf{d}(\mathbf{u});
    \mathbf{m}(\mathbf{u})/s;
    \mathbf{o}/s
    \right] \in \mathbb{R}^{9},
\end{equation}
where $s$ is a fixed scene-scale normalization factor.
We construct the camera map at the latent spatial resolution, broadcasting the camera center over spatial locations.
The resulting map is concatenated channel-wise with the inputs to both the RGB and pointmap streams before patch embedding, providing viewpoint information for generation.
Note that in tabletop robotic manipulation, external cameras are often fixed, whereas wrist-mounted cameras move with the robot.
Our formulation supports conditioning on different camera poses, although modeling camera motion is not a primary focus of this work.

\vspace{-0.3cm}
\paragraph{Condition Dropout.}
To support flexible inference configurations, we independently apply action and camera dropout during training, with an additional joint text-action dropout strategy.
Missing conditions are represented by zero embeddings, allowing the model to generate future observations with different combinations of available conditions.
The initial RGB observations and pointmaps are always retained, while the language instruction may be dropped jointly with the action condition during training.

\subsection{Overall Training Objective}
\label{sec:training_objective}

The RGB and pointmap streams are trained with separate latent flow-matching losses, $\mathcal{L}^{\mathrm{RGB}}_{\mathrm{FM}}$ and $\mathcal{L}^{\mathrm{P}}_{\mathrm{FM}}$, which supervise velocity prediction in their respective latent spaces.
We combine these losses with the two cross-view supervision objectives to obtain
\begin{equation}
\begin{aligned}
    \mathcal{L}
    =\mathcal{L}^{\mathrm{RGB}}_{\mathrm{FM}}
    + \mathcal{L}^{\mathrm{P}}_{\mathrm{FM}}
    + \lambda_{\mathrm{point}}(k)
    \mathcal{L}_{\mathrm{point}}
    + \lambda_{\mathrm{feat}}(k)
    \mathcal{L}_{\mathrm{feat}},
\end{aligned}
\end{equation}
where $\mathcal{L}_{\mathrm{point}}$ and $\mathcal{L}_{\mathrm{feat}}$ denote the cross-view pointmap consistency and geometry-aware feature alignment losses, respectively.
The weights $\lambda_{\mathrm{point}}(k)$ and $\lambda_{\mathrm{feat}}(k)$ vary with training step $k$.
We linearly warm up each auxiliary term and use detached loss-magnitude rescaling to bound its contribution relative to the combined flow-matching losses, preventing cross-view supervision from dominating the generation objective.

\begin{table*}[t]
\centering
\caption{Comparison on real-world robotic manipulation tasks.}
\label{tab:sota_comparison_real}
\small
\setlength{\tabcolsep}{4pt}
\resizebox{0.9\textwidth}{!}{%
\begin{tabular}{l cc cccc cc}
\toprule
\multirow{2}{*}{Method}
& \multicolumn{2}{c}{RGB Video Quality}
& \multicolumn{4}{c}{Depth Accuracy}
& \multicolumn{2}{c}{Cross-View Consistency} \\
\cmidrule(lr){2-3}
\cmidrule(lr){4-7}
\cmidrule(lr){8-9}
& $\mathrm{FVD}_{n}$ $\downarrow$
& $\mathrm{FVD}_{m}$ $\downarrow$
& $\mathrm{Abs\,Rel}_{n}$ $\downarrow$
& $\mathrm{Abs\,Rel}_{m}$ $\downarrow$
& $\delta_{1,n}$ $\uparrow$
& $\delta_{1,m}$ $\uparrow$
& CD $\downarrow$
& mIoU $\uparrow$ \\
\midrule
Wan2.1\cite{wan2025wan}
& 333.5 & 323.8 & -- & -- & -- & -- & -- & -- \\
SVD\cite{blattmann2023stable}
& 634.2 & 637.5 & -- & -- & -- & -- & -- & -- \\
4DGen\cite{liu2026geometry}
& 457.5 & 475.3 & 0.065 & 0.056 & 0.944 & 0.948 & 0.028 & 0.452 \\
TesserAct\cite{zhen2025tesseract}
& 783.9 & 821.2 & 0.098 & 0.081 & 0.917 & 0.925 & -- & -- \\
Geometry Forcing\cite{wu2026geometry}
& 1251.1 & 1245.1 & 0.198 & 0.174 & 0.527 & 0.612 & -- & -- \\
\midrule
\textbf{Ours}
& \textbf{259.1} & \textbf{258.4} & \textbf{0.027} & \textbf{0.024} & \textbf{0.978} & \textbf{0.977} & \textbf{0.011} & \textbf{0.461} \\
\bottomrule
\end{tabular}%
}
\end{table*}

\begin{figure*}[t!]
\centering
\includegraphics[
  width=0.92\textwidth
]{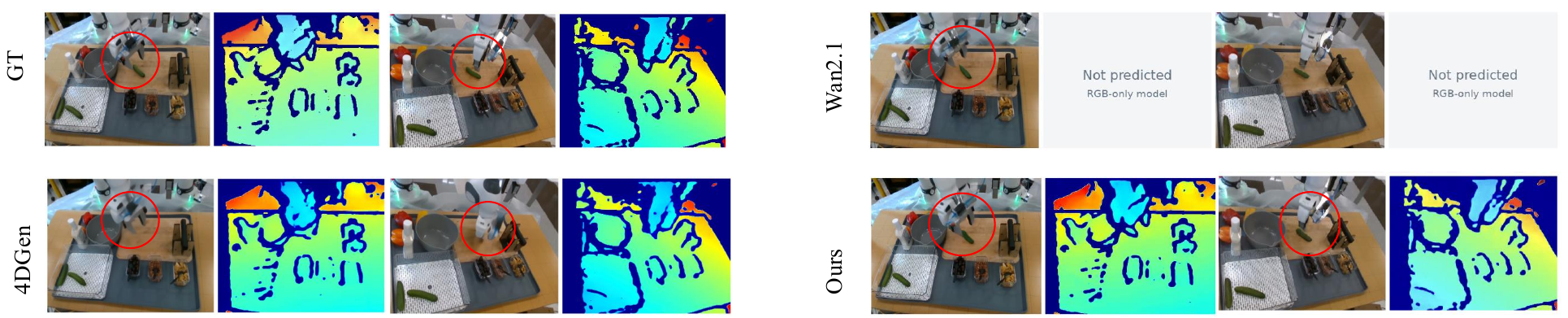}
\caption{Qualitative comparison with state-of-the-art methods on real-world robotic manipulation tasks.}
\label{fig:qualitative_real}
\end{figure*}

\section{Experiments}

\subsection{Experimental Setup}

\paragraph{Dataset.}
We evaluate GeoMVGen on the 4DGen~\cite{liu2026geometry} dataset, comprising three simulated robotic manipulation tasks and four real-world tasks.
Each sample contains synchronized observations from two viewpoints.
For simulated tasks, we train on 12 camera viewpoints and evaluate on four held-out viewpoints to assess generalization to unseen camera perspectives.
For real-world tasks, we use two fixed cameras and the same 40 episodes for training and evaluation, focusing on in-domain generation.
All methods are evaluated on identical clips and camera pairs.

\vspace{-0.3cm}
\paragraph{Evaluation Metrics.}
RGB video quality is evaluated using Fr\'echet Video Distance (FVD)~\cite{liu2024fr}.
Depth accuracy is measured using absolute relative error (Abs Rel) and threshold accuracy ($\delta_1$).
For cross-view consistency, we report Chamfer Distance (CD, in meters) to measure geometric agreement between predicted point clouds across views and mean Intersection-over-Union (mIoU) to assess cross-view gripper alignment.
Subscripts $n$ and $m$ denote the reference and secondary views, respectively.
Details of the evaluation protocols are provided in the supplementary material.

\vspace{-0.3cm}
\paragraph{Baselines.}
We compare GeoMVGen against three categories of methods: general-purpose video generators, including Wan2.1~\cite{wan2025wan} and Stable Video Diffusion (SVD)~\cite{blattmann2023stable};
joint appearance-geometry generation models, including TesserAct~\cite{zhen2025tesseract} and 4DGen~\cite{liu2026geometry};
and the geometry foundation model-guided approach Geometry Forcing~\cite{wu2026geometry}.

\subsection{Comparison with State-of-the-Art Methods}

\paragraph{Quantitative Comparison.}
Table~\ref{tab:sota_comparison} reports the quantitative results on simulated robotic manipulation tasks, where GeoMVGen achieves the best performance across all reported metrics.
For RGB video quality, GeoMVGen achieves the lowest FVD scores of 480.2 and 486.5, outperforming general-purpose video generators and joint appearance-geometry baselines.
For depth accuracy, our method achieves the lowest Abs Rel of 0.051 and 0.049 and the highest $\delta_1$ of 0.957 and 0.960.
For cross-view consistency, compared with 4DGen, GeoMVGen reduces Chamfer distance from 0.029 to 0.013 and improves mIoU from 0.496 to 0.530.
Table~\ref{tab:sota_comparison_real} shows similar improvements on real-world robotic manipulation tasks.
Our method achieves the lowest FVD scores of 259.1 and 258.4, the lowest Abs Rel of 0.027 and 0.024, and the highest $\delta_1$ of 0.978 and 0.977.
Compared with 4DGen, GeoMVGen further reduces Chamfer distance from 0.028 to 0.011 and improves cross-view mIoU from 0.452 to 0.461.
These results demonstrate the effectiveness of GeoMVGen in jointly generating realistic appearance and accurate geometry while maintaining consistency across viewpoints in both simulated and real-world robotic manipulation scenarios.

\vspace{-0.3cm}
\paragraph{Qualitative Comparison.}
Figures~\ref{fig:qualitative} and~\ref{fig:qualitative_real} present qualitative comparisons on simulated and real-world robotic manipulation tasks, respectively.
In simulated tasks, GeoMVGen maintains more consistent object appearance across views, whereas 4DGen shows inconsistent apple colors in the second viewpoint.
In real-world scenarios, 4DGen occasionally fails to preserve the interacting object in generated frames, while GeoMVGen maintains its visibility.
The generated depth maps further reflect more accurate geometric predictions, including the spatial relationships between the gripper and manipulated objects.
GeoMVGen also produces more accurate gripper positioning relative to the ground truth and better preserves spatial coherence across views.
These comparisons illustrate the benefits of joint appearance-geometry generation for capturing robot-object interactions across viewpoints.

\begin{table}[t!]
\centering
\caption{{Ablation of Joint Appearance-Geometry Modeling Design.} For modality-specific adaptation, \cmark~denotes separate adaptation modules, while \xmark~denotes shared adaptation modules. RGB and depth metrics are reported for the reference view.}
\label{tab:cross_modal_ablation}
\small
\setlength{\tabcolsep}{4pt}
\resizebox{0.49\textwidth}{!}{%
\begin{tabular}{cc c cc c}
\toprule
\multirow{2}{*}{\shortstack{Modal-specific\\adaptation}}
& \multirow{2}{*}{\shortstack{Cross-modal\\residual}}
& \multicolumn{1}{c}{RGB}
& \multicolumn{2}{c}{Depth}
& \multicolumn{1}{c}{Cross-View} \\
\cmidrule(lr){3-3}
\cmidrule(lr){4-5}
\cmidrule(lr){6-6}
& & $\mathrm{FVD}_{n}$ $\downarrow$
& $\mathrm{AbsRel}_{n}$ $\downarrow$
& $\delta_{1,n}$ $\uparrow$
& mIoU $\uparrow$ \\
\midrule
\xmark & \xmark
& 663.3 & 0.049 & 0.953 & 0.473 \\
\cmark & \xmark
& 643.2 & 0.050 & 0.953 & 0.485 \\
\xmark & \cmark
& 587.4 & 0.047 & 0.957 & 0.494 \\
\cmark & \cmark
& \textbf{566.0} & \textbf{0.045} & \textbf{0.959} & \textbf{0.504} \\
\bottomrule
\end{tabular}%
}
\end{table}

\begin{figure*}[t!]
\centering
\begin{tikzpicture}
\begin{axis}[ybar, width=0.3\textwidth, height=5cm, bar width=12pt, ymin=29, ymax=30.8, ylabel={ADE (pixels)}, xtick={1,2,3,4}, xticklabels={Wan2.1,SVD,4DGen,Ours}, xmin=0.5, xmax=4.5, tick label style={font=\scriptsize}, label style={font=\small}, nodes near coords, every node near coord/.append style={ font=\tiny, anchor=south }, point meta=y, scaled y ticks=false,]
\addplot[ybar, bar shift=0pt, fill=gray!55, draw=none] coordinates {
    (1,30.41)
    (2,29.93)
    (3,29.84)
};
\addplot[ybar, bar shift=0pt, fill=red!75, draw=none] coordinates {
    (4,29.55)
};
\end{axis}
\end{tikzpicture}
\hfill
\begin{tikzpicture}
\begin{axis}[ybar, width=0.3\textwidth, height=5cm, bar width=12pt, ymin=40, ymax=42, ylabel={FDE (pixels)}, xtick={1,2,3,4}, xticklabels={Wan2.1,SVD,4DGen,Ours}, xmin=0.5, xmax=4.5, tick label style={font=\scriptsize}, label style={font=\small}, nodes near coords, every node near coord/.append style={ font=\tiny, anchor=south }, point meta=y, scaled y ticks=false,]
\addplot[ybar, bar shift=0pt, fill=gray!55, draw=none] coordinates {
    (1,41.69)
    (2,41.16)
    (3,40.80)
};
\addplot[ybar, bar shift=0pt, fill=red!75, draw=none] coordinates {
    (4,40.55)
};
\end{axis}
\end{tikzpicture}
\hfill
\begin{tikzpicture}
\begin{axis}[ybar, width=0.3\textwidth, height=5cm, bar width=22pt, ymin=0, ymax=1.3, ylabel={Error reduction (\%)}, xtick={1,2}, xticklabels={ADE,FDE}, xmin=0.4, xmax=2.6, tick label style={font=\scriptsize}, label style={font=\small}, nodes near coords, every node near coord/.append style={ font=\small, anchor=south, /pgf/number format/fixed, /pgf/number format/precision=2 }, point meta=y, scaled y ticks=false,]
\addplot[fill=red!75, draw=none] coordinates {
    (1,0.97)
    (2,0.61)
};
\end{axis}
\end{tikzpicture}
\caption{{Average gripper trajectory accuracy across three simulation tasks.} (a) Average displacement error (ADE). (b) Final displacement error (FDE). (c) Relative error reduction of our method compared with the strongest baseline, 4DGen. Lower ADE and FDE indicate better performance. The ADE and FDE axes are truncated for visibility.}
\label{fig:gripper_trajectory}
\end{figure*}
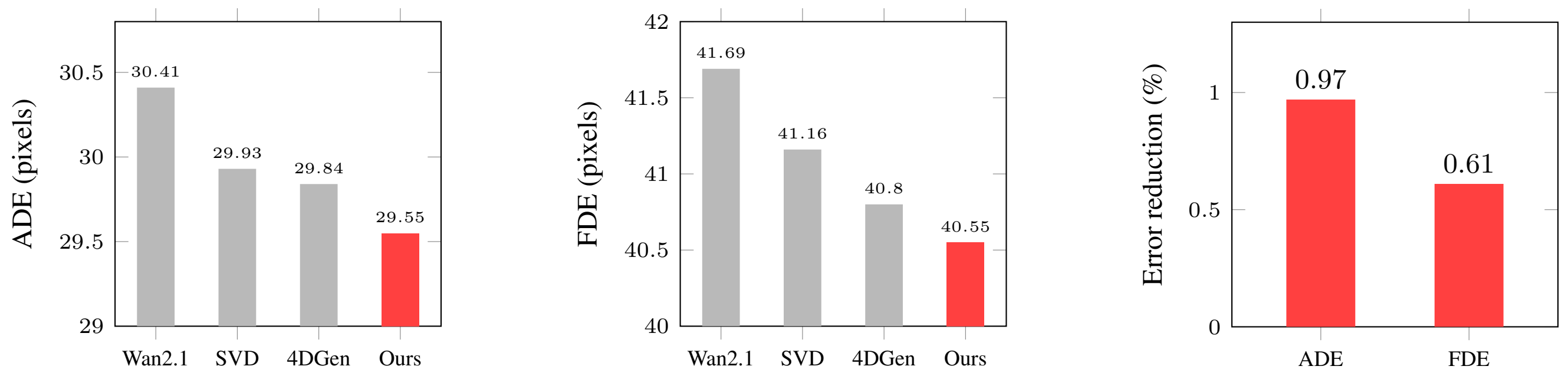
\subsection{Ablation Studies}

In this section, we conduct all ablation experiments on the \textit{PlaceAppleFromBowlIntoBin} task with 1,000 training steps per configuration due to computational constraints but ensure fair comparisons across variants.
Within each ablation group, all experimental settings are kept identical except for the component under investigation.

\vspace{-0.3cm}
\paragraph{Ablation of Joint Appearance-Geometry Modeling Design.}
Table~\ref{tab:cross_modal_ablation} evaluates the contributions of modality-specific adaptation and bidirectional cross-modal residual connections.
Introducing cross-modal residual connections consistently improves RGB video quality, depth accuracy, and cross-view consistency under both shared and modality-specific adaptation.
Modality-specific adaptation further improves RGB video quality and cross-view consistency, while its effects on depth accuracy vary without cross-modal interaction.
Combining both components achieves the best performance across all reported metrics, with the lowest FVD (566.0) and Abs Rel (0.045), and the highest $\delta_1$ (0.959) and mIoU (0.504).
These results validate the complementary benefits of specialized modality representations and bidirectional information exchange for joint appearance-geometry generation.

\begin{table}[t]
\centering
\caption{{Ablation of Cross-View Consistency Mechanism.} Pointmap consistency denotes cross-view pointmap consistency supervision, and feature alignment denotes geometry-aware cross-view feature alignment. Joint supervision combines both objectives. RGB and depth metrics are reported for the reference view.}
\label{tab:cross_view_ablation}
\small
\setlength{\tabcolsep}{5pt}
\resizebox{0.49\textwidth}{!}{%
\begin{tabular}{l c cc cc}
\toprule
\multirow{2}{*}{Supervision}
& \multicolumn{1}{c}{RGB}
& \multicolumn{2}{c}{Depth}
& \multicolumn{2}{c}{Cross-View} \\
\cmidrule(lr){2-2}
\cmidrule(lr){3-4}
\cmidrule(lr){5-6}
& $\mathrm{FVD}_{n}$ $\downarrow$
& $\mathrm{AbsRel}_{n}$ $\downarrow$
& $\delta_{1,n}$ $\uparrow$
& CD (m) $\downarrow$
& mIoU $\uparrow$ \\
\midrule
No cross-view supervision
& 550.0 & 0.047 & \textbf{0.960} & 0.0163 & 0.496 \\
Pointmap consistency
& 532.4 & 0.048 & \textbf{0.960} & 0.0160 & \textbf{0.508} \\
Feature alignment
& 527.7 & \textbf{0.045} & \textbf{0.960} & 0.0162 & 0.481 \\
Joint supervision
& \textbf{506.7} & 0.051 & \textbf{0.960} & \textbf{0.0159} & 0.501 \\
\bottomrule
\end{tabular}%
}
\end{table}

\vspace{-0.3cm}
\paragraph{Ablation of Cross-View Consistency Mechanism.}
Table~\ref{tab:cross_view_ablation} investigates the contributions of cross-view pointmap consistency and geometry-aware feature alignment.
Both objectives individually improve RGB video quality, while their combination achieves the lowest FVD (506.7).
For depth accuracy, feature alignment achieves the lowest Abs Rel (0.045), whereas joint supervision exhibits a trade-off in depth accuracy, with $\delta_1$ remaining unchanged across all configurations.
For cross-view consistency, pointmap consistency achieves the highest mIoU (0.508), while joint supervision achieves the lowest CD (0.0159 m) and improves mIoU over the model without cross-view supervision.
These results demonstrate that the two objectives provide complementary geometric and feature-level constraints, improving RGB video quality and cross-view geometric consistency despite trade-offs in individual metrics.

\subsection{Gripper Trajectory Accuracy Evaluation}

To evaluate how accurately generated videos capture task-relevant gripper motion and final positioning, we measure gripper trajectory accuracy across three simulated manipulation tasks.
All methods are evaluated on matched clips and camera pairs.
A shared SAM2~\cite{ravi2025sam}-based tracker is initialized with ground-truth gripper masks in the conditioning frame and tracks each generated RGB view without future ground-truth prompts.
We compare the tracked mask centroids with ground-truth gripper trajectories over ten future frames.
Average displacement error (ADE) measures the mean positional error across frames.
Final displacement error (FDE) measures the error at the final frame.
Both metrics are reported in pixels over valid ground-truth targets.

\vspace{-0.3cm}
\paragraph{Results.}
Figure~\ref{fig:gripper_trajectory} presents the average gripper trajectory errors across the three simulated manipulation tasks.
GeoMVGen achieves the lowest ADE (29.55 pixels) and FDE (40.55 pixels) among the compared methods, reducing these errors by 0.97\% and 0.61\% relative to 4DGen, respectively.
Although modest, these improvements indicate that our method more accurately captures task-relevant gripper motion and final positioning in generated videos.
\section{Conclusion}
\label{sec:conclusion}

This work presents GeoMVGen, a new framework for joint multi-view appearance-geometry generation in robotic manipulation.
By combining modality-specific adaptation with cross-modal connections, our framework enables effective joint modeling of appearance and geometry.
Complementary cross-view supervision promotes geometric consistency, while optional action and camera conditioning supports flexible generation.
Extensive experiments on simulated and real-world robotic manipulation tasks demonstrate improvements in video quality, geometric accuracy, and cross-view consistency over state-of-the-art methods.

{
    \small
    \bibliographystyle{ieeenat_fullname}
    \bibliography{main}

@article{blattmann2023stable,
  title={Stable video diffusion: Scaling latent video diffusion models to large datasets},
  author={Blattmann, Andreas and Dockhorn, Tim and Kulal, Sumith and Mendelevitch, Daniel and Kilian, Maciej and Lorenz, Dominik and Levi, Yam and English, Zion and Voleti, Vikram and Letts, Adam and others},
  journal={arXiv preprint arXiv:2311.15127},
  year={2023}
}

@inproceedings{yang2025cogvideox,
  title={Cogvideox: Text-to-video diffusion models with an expert transformer},
  author={Yang, Zhuoyi and Teng, Jiayan and Zheng, Wendi and Ding, Ming and Huang, Shiyu and Xu, Jiazheng and Yang, Yuanming and Hong, Wenyi and Zhang, Xiaohan and Feng, Guanyu and others},
  booktitle={International Conference on Learning Representations},
  volume={2025},
  pages={83048--83077},
  year={2025}
}

@article{wan2025wan,
  title={Wan: Open and advanced large-scale video generative models},
  author={Wan, Team and Wang, Ang and Ai, Baole and Wen, Bin and Mao, Chaojie and Xie, Chen-Wei and Chen, Di and Yu, Feiwu and Zhao, Haiming and Yang, Jianxiao and others},
  journal={arXiv preprint arXiv:2503.20314},
  year={2025}
}

@inproceedings{chen2024videocrafter2,
  title={Videocrafter2: Overcoming data limitations for high-quality video diffusion models},
  author={Chen, Haoxin and Zhang, Yong and Cun, Xiaodong and Xia, Menghan and Wang, Xintao and Weng, Chao and Shan, Ying},
  booktitle={2024 IEEE/CVF Conference on Computer Vision and Pattern Recognition (CVPR)},
  pages={7310--7320},
  year={2024},
  organization={IEEE}
}

@inproceedings{bar2024lumiere,
  title={Lumiere: A space-time diffusion model for video generation},
  author={Bar-Tal, Omer and Chefer, Hila and Tov, Omer and Herrmann, Charles and Paiss, Roni and Zada, Shiran and Ephrat, Ariel and Hur, Junhwa and Liu, Guanghui and Raj, Amit and others},
  booktitle={SIGGRAPH Asia 2024 conference papers},
  pages={1--11},
  year={2024}
}

@article{kong2024hunyuanvideo,
  title={Hunyuanvideo: A systematic framework for large video generative models},
  author={Kong, Weijie and Tian, Qi and Zhang, Zijian and Min, Rox and Dai, Zuozhuo and Zhou, Jin and Xiong, Jiangfeng and Li, Xin and Wu, Bo and Zhang, Jianwei and others},
  journal={arXiv preprint arXiv:2412.03603},
  year={2024}
}

@article{du2023learning,
  title={Learning universal policies via text-guided video generation},
  author={Du, Yilun and Yang, Sherry and Dai, Bo and Dai, Hanjun and Nachum, Ofir and Tenenbaum, Josh and Schuurmans, Dale and Abbeel, Pieter},
  journal={Advances in neural information processing systems},
  volume={36},
  pages={9156--9172},
  year={2023}
}

@article{hu2024video,
  title={Video prediction policy: A generalist robot policy with predictive visual representations},
  author={Hu, Yucheng and Guo, Yanjiang and Wang, Pengchao and Chen, Xiaoyu and Wang, Yen-Jen and Zhang, Jianke and Sreenath, Koushil and Lu, Chaochao and Chen, Jianyu},
  journal={arXiv preprint arXiv:2412.14803},
  year={2024}
}

@article{liang2025video,
  title={Video generators are robot policies},
  author={Liang, Junbang and Tokmakov, Pavel and Liu, Ruoshi and Sudhakar, Sruthi and Shah, Paarth and Ambrus, Rares and Vondrick, Carl},
  journal={arXiv preprint arXiv:2508.00795},
  year={2025}
}

@article{bharadhwaj2024gen2act,
  title={Gen2act: Human video generation in novel scenarios enables generalizable robot manipulation},
  author={Bharadhwaj, Homanga and Dwibedi, Debidatta and Gupta, Abhinav and Tulsiani, Shubham and Doersch, Carl and Xiao, Ted and Shah, Dhruv and Xia, Fei and Sadigh, Dorsa and Kirmani, Sean},
  journal={arXiv preprint arXiv:2409.16283},
  year={2024}
}

@article{zhen2025tesseract,
  title={Tesseract: learning 4d embodied world models},
  author={Zhen, Haoyu and Sun, Qiao and Zhang, Hongxin and Li, Junyan and Zhou, Siyuan and Du, Yilun and Gan, Chuang},
  journal={arXiv preprint arXiv:2504.20995},
  year={2025}
}

@article{chen20254dnex,
  title={4dnex: Feed-forward 4d generative modeling made easy},
  author={Chen, Zhaoxi and Liu, Tianqi and Zhuo, Long and Ren, Jiawei and Tao, Zeng and Zhu, He and Hong, Fangzhou and Pan, Liang and Liu, Ziwei},
  journal={arXiv preprint arXiv:2508.13154},
  year={2025}
}

@inproceedings{mi2026one4d,
  title={One4d: Unified 4d generation and reconstruction via decoupled lora control},
  author={Mi, Zhenxing and Wang, Yuxin and Xu, Dan},
  booktitle={European Conference on Computer Vision},
  pages={38--56},
  year={2026},
  organization={Springer}
}

@article{chen2026unividx,
  title={UniVidX: A Unified Multimodal Framework for Versatile Video Generation via Diffusion Priors},
  author={Chen, Houyuan and Li, Hong and Kong, Xianghao and Zhu, Tianrui and Xu, Shaocong and Xiao, Weiqing and Guo, Yuwei and Ye, Chongjie and Zhang, Lvmin and Zhao, Hao and others},
  journal={arXiv preprint arXiv:2605.00658},
  year={2026}
}

@inproceedings{xie2025sv4d,
  title={Sv4d: Dynamic 3d content generation with multi-frame and multi-view consistency},
  author={Xie, Yiming and Yao, Chun-Han and Voleti, Vikram and Jiang, Huaizu and Jampani, Varun},
  booktitle={International Conference on Learning Representations},
  volume={2025},
  pages={33421--33441},
  year={2025}
}

@inproceedings{wu2025cat4d,
  title={Cat4d: Create anything in 4d with multi-view video diffusion models},
  author={Wu, Rundi and Gao, Ruiqi and Poole, Ben and Trevithick, Alex and Zheng, Changxi and Barron, Jonathan T and Holynski, Aleksander},
  booktitle={Proceedings of the IEEE/CVF Conference on Computer Vision and Pattern Recognition},
  pages={26057--26068},
  year={2025}
}

@inproceedings{liu2026geometry,
  title={Geometry-aware 4d video generation for robot manipulation},
  author={Liu, Zeyi and Li, Shuang and Cousineau, Eric and Feng, Siyuan and Burchfiel, Benjamin and Song, Shuran},
  booktitle={International Conference on Learning Representations},
  volume={2026},
  pages={140751--140773},
  year={2026}
}

@article{wang2026mvista,
  title={MVISTA-4D: View-consistent 4d world model with test-time action inference for robotic manipulation},
  author={Wang, Jiaxu and Jiang, Yicheng and He, Tianlun and Sun, Jingkai and Zhang, Qiang and Cao, Jiahang and Gan, Zesen and Sun, Mingyuan and Shao, Qiming and Yue, Xiangyu},
  journal={arXiv preprint arXiv:2602.09878},
  year={2026}
}

@article{kuang2024collaborative,
  title={Collaborative video diffusion: Consistent multi-video generation with camera control},
  author={Kuang, Zhengfei and Cai, Shengqu and He, Hao and Xu, Yinghao and Li, Hongsheng and Guibas, Leonidas J and Wetzstein, Gordon},
  journal={Advances in Neural Information Processing Systems},
  volume={37},
  pages={16240--16271},
  year={2024}
}

@inproceedings{bai2025syncammaster,
  title={Syncammaster: Synchronizing multi-camera video generation from diverse viewpoints},
  author={Bai, Jianhong and Xia, Menghan and Wang, Xintao and Yuan, Ziyang and Liu, Zuozhu and Hu, Haoji and Wan, Pengfei and Zhang, Di},
  booktitle={International Conference on Learning Representations},
  volume={2025},
  pages={58038--58060},
  year={2025}
}

@inproceedings{wang2025vggt,
    author    = {Wang, Jianyuan and Chen, Minghao and Karaev, Nikita and Vedaldi, Andrea and Rupprecht, Christian and Novotny, David},
    title     = {{VGGT}: Visual Geometry Grounded Transformer},
    booktitle = {Proceedings of the Computer Vision and Pattern Recognition Conference (CVPR)},
    month     = {June},
    year      = {2025},
    pages     = {5294-5306}
}

@InProceedings{wang2026vggt,
    author    = {Wang, Jianyuan and Chen, Minghao and Zhang, Shangzhan and Karaev, Nikita and Sch\"onberger, Johannes and Labatut, Patrick and Bojanowski, Piotr and Novotny, David and Vedaldi, Andrea and Rupprecht, Christian},
    title     = {{VGGT-$\Omega$}},
    booktitle = {Proceedings of the IEEE/CVF Conference on Computer Vision and Pattern Recognition (CVPR)},
    month     = {June},
    year      = {2026},
    pages     = {21486-21499}
}

@inproceedings{wu2026geometry,
  title={Geometry forcing: Marrying video diffusion and 3d representation for consistent world modeling},
  author={Wu, Haoyu and Wu, Diankun and He, Tianyu and Guo, Junliang and Ye, Yang and Duan, Yueqi and Bian, Jiang},
  booktitle={International Conference on Learning Representations},
  volume={2026},
  pages={119186--119209},
  year={2026}
}

@InProceedings{Wu_2026_CVPR,
    author    = {Wu, Guile and Huang, David and Bai, Dongfeng and Liu, Bingbing},
    title     = {{MoVieDrive}: Urban Scene Synthesis with Multi-Modal Multi-View Video Diffusion Transformer},
    booktitle = {Proceedings of the IEEE/CVF Conference on Computer Vision and Pattern Recognition (CVPR) Findings},
    month     = {June},
    year      = {2026},
    pages     = {4290-4299}
}

@article{danier2025view,
  title={View-consistent diffusion representations for 3d-consistent video generation},
  author={Danier, Duolikun and Gao, Ge and McDonagh, Steven and Li, Changjian and Bilen, Hakan and Mac Aodha, Oisin},
  journal={arXiv preprint arXiv:2511.18991},
  year={2025}
}

@article{hu2021lora,
  title={Lora: Low-rank adaptation of large language models},
  author={Hu, Edward J and Shen, Yelong and Wallis, Phillip and Allen-Zhu, Zeyuan and Li, Yuanzhi and Wang, Shean and Wang, Lu and Chen, Weizhu},
  journal={arXiv preprint arXiv:2106.09685},
  year={2021}
}

@inproceedings{ravi2025sam,
  title={Sam 2: Segment anything in images and videos},
  author={Ravi, Nikhila and Gabeur, Valentin and Hu, Yuan-Ting and Hu, Ronghang and Ryali, Chaitanya and Ma, Tengyu and Khedr, Haitham and R{\"a}dle, Roman and Rolland, Chloe and Gustafson, Laura and others},
  booktitle={International Conference on Learning Representations},
  volume={2025},
  pages={28085--28128},
  year={2025}
}

@article{liu2024fr,
  title={Fr$\backslash$'echet Video Motion Distance: A Metric for Evaluating Motion Consistency in Videos},
  author={Liu, Jiahe and Qu, Youran and Yan, Qi and Zeng, Xiaohui and Wang, Lele and Liao, Renjie},
  journal={arXiv preprint arXiv:2407.16124},
  year={2024}
}
}

\end{document}